\documentclass[journal]{IEEEtran}
\newcommand{\journalname}{IEEE Transactions on Medical Imaging}
\usepackage{cite}
\usepackage{amsmath,amssymb,amsfonts}
\usepackage{algorithmic}
\usepackage{graphicx}
\usepackage{textcomp}
\usepackage{dsfont}
\usepackage{amssymb}
\usepackage{multirow}
\usepackage{url}
\usepackage{colortbl}
\usepackage{xcolor}
\definecolor{lightergray}{gray}{0.95}
\def\BibTeX{{\rm B\kern-.05em{\sc i\kern-.025em b}\kern-.08em
    T\kern-.1667em\lower.7ex\hbox{E}\kern-.125emX}}
\begin{document}

\title{Double Banking on Knowledge: Customized Modulation and Prototypes for Multi-Modality Semi-supervised Medical Image Segmentation}

\author{Yingyu Chen, Ziyuan Yang, Ming Yan, Zhongzhou Zhang, Hui Yu, Yan Liu, Yi Zhang, \IEEEmembership{Senior Member, IEEE}
\thanks{Y. Chen, Z. Yang, Z. Zhang, and H. Yu are with the College of Computer Science, Sichuan University, Chengdu 610065, China (e-mail: cyy262511@gmail.com, cziyuanyang@gmail.com, zz\_zhang@stu.scu.edu.cn, smileeudora@163.com).}
\thanks{M. Yan is with the Agency for Science, Technology and Research (A*STAR), Singapore (e-mail: yanmingtop@gmail.com).}
\thanks{Y. Liu is with the College of Electrical Engineering, Sichuan University, Chengdu 610065, China (e-mail: liuyan77@scu.edu.cn).}
\thanks{Y. Zhang is with the School of Cyber Science and Engineering, Sichuan University, Chengdu 610065, China (e-mail: yzhang@scu.edu.cn).}
\thanks{Corresponding author: Yi Zhang.}
}

\maketitle
\begin{abstract}
Multi-modality (MM) semi-supervised learning (SSL) based medical image segmentation has recently gained increasing attention for its ability to utilize MM data and reduce reliance on labeled images. However, current methods face several challenges: 
(1) Complex network designs hinder scalability to scenarios with more than two modalities. 
(2) Focusing solely on modality-invariant representation while neglecting modality-specific features, leads to incomplete MM learning.
(3) Leveraging unlabeled data with generative methods can be unreliable for SSL. 
To address these problems, we propose Double Bank Dual Consistency (DBDC), a novel MM-SSL approach for medical image segmentation. 
To address challenge (1), we propose a modality all-in-one segmentation network that accommodates data from any number of modalities, removing the limitation on modality count.
To address challenge (2), we design two learnable plug-in banks, Modality-Level Modulation bank (MLMB) and Modality-Level Prototype (MLPB) bank, to capture both modality-invariant and modality-specific knowledge. These banks are updated using our proposed Modality Prototype Contrastive Learning (MPCL).
Additionally, we design Modality Adaptive Weighting (MAW) to dynamically adjust learning weights for each modality, ensuring balanced MM learning as different modalities learn at different rates. Finally, to address challenge (3), we introduce a Dual Consistency (DC) strategy that enforces consistency at both the image and feature levels without relying on generative methods.
We evaluate our method on a 2-to-4 modality segmentation task using three open-source datasets, and extensive experiments show that our method outperforms state-of-the-art approaches.

\end{abstract}


\section{Introduction}
\label{sec:introduction}
\IEEEPARstart{M}{ulti}-modality imaging techniques are widely employed in clinical diagnosis since different imaging modalities provide distinct advantages in contrast and anatomical detail visualization. 
Recently, several deep learning-based MM medical image segmentation methods have achieved significant success~\cite{douqi, MASS,semi-CML}, which plays a vital role in clinical decision-making.
By leveraging MM information, these methods generally outperform traditional single-modality image segmentation techniques~\cite{MASS,douqi,SUMML,semi-CML}.

\begin{figure}[]
    \centering
    \includegraphics[width=0.35\textwidth]{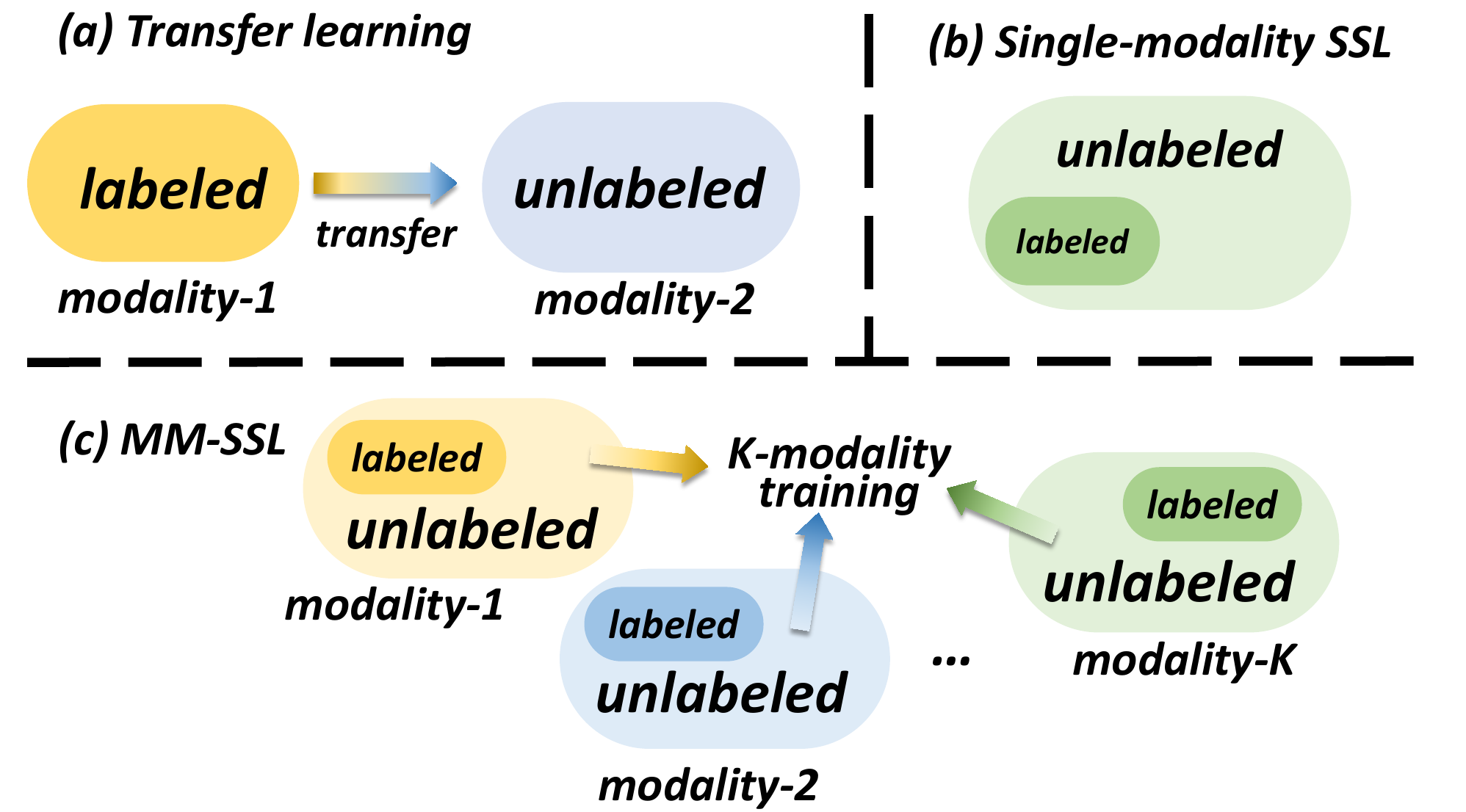}
    \caption{(a) Transfer learning task. (b)  Single modality SSL task. (c) MM-SSL task.}
    \label{fig1}
    \vspace{-15pt}
\end{figure}

However, training an effective model requires a large amount of well-labeled data, and obtaining segmentation annotations is a time-consuming and labor-intensive process that requires specialized medical expertise \cite{MeanTeacher}. This challenge is further intensified in MM scenarios, where the complexity and cost of annotations are considerably higher \cite{MASS}. Transfer learning is a viable approach to reduce dependency on labeled data by utilizing fully labeled source modality data and applying domain adaptation or generalization techniques to segment unlabeled target modality data~\cite{RAM-DSIR, VarDA}, as illustrated in Fig.~\ref{fig1}(a).
However, transfer learning methods still heavily rely on a large amount of well-labeled source modality data, and their performance is sensitive to domain gaps. In scenarios with substantial modality gaps (e.g., between computed tomography (CT) and magnetic resonance (MR)), these methods often fail to keep their performance~\cite{causal}.

Semi-supervised learning (SSL) offers a promising approach that enables training with a small amount of labeled data alongside a large amount of unlabeled data. However, current mainstream SSL methods focus on single-modality scenarios and lack generalization to MM medical imaging, as shown in Fig.~\ref{fig1}(b). To address this limitation, multi-modality SSL (MM-SSL) has recently garnered significant attention, as it combines the characteristics of MM learning scenarios with the low data dependency of SSL. This task leverages MM data, where each modality has a small amount of labeled data and a large amount of unlabeled data, as shown in Fig.~\ref{fig1}(c).
Compared to single-modality SSL, MM-SSL is designed for MM scenarios and offers better generalization, maintaining performance across different modalities.
Besides, compared to transfer learning, it requires less labeled data and does not need finetuning in the target domain.

Current MM-SSL medical image segmentation methods can be classified into two categories: (1) Using \textbf{registered data}, where different modalities share the same segmentation mask~\cite{semi-CML}. However, obtaining registered data is highly challenging, as it requires collecting all modalities from the same patient, which is quite difficult~\cite{zhang2023multi}.
(2) Using \textbf{unregistered data}, where different modalities have separate segmentation masks. This data is easier to obtain, as it only requires gathering diverse modality datasets from different patients. Therefore, approaches using unregistered data offer better generalization and practical value. For instance, Zhu \textit{et al.}~\cite{SUMML} proposed SUMML, a dual-modality SSL method using CycleGAN~\cite{CycleGAN} to generate data from one modality to the other, thereby expanding the labeled dataset. 
Chen \textit{et al.}~\cite{MASS} introduced MASS, a 2-modality SSL that uses modality-specific encoders to extract modality-invariant features and a shared decoder to generate registration matrices. During inference, the network processes both labeled and unlabeled data to generate corresponding registration matrices. The registration matrices are applied to predict the segmentation of unlabeled data.

Nevertheless, these MM-SSL methods face several challenges.
\textbf{\textit{(i)}} These methods are tailored for 2-modality scenarios and rely on multiple independent networks or complex networks, making it difficult to extend them to more modalities. For example, SUMML assigns different segmentation and generative networks for each modality and generates pseudo-data to expand the training set. MASS requires a modality registration operation~\cite{warp}, specifically designed for 2-modality scenarios. As the number of modalities increases, the complexity of these methods grows exponentially. 
\textbf{\textit{(ii)}} Incomplete MM learning. For instance, MASS uses a shared decoder for learning modality-invariant features, but it neglects modality-specific feature representation.
\textbf{\textit{(iii)}} The performance of SUMML heavily depends on the quality of images generated by CycleGAN, which is difficult to control during training. MASS relies on a learnable registration network~\cite{warp} for unlabeled data learning, but its performance is highly dependent on registration accuracy. The data-intensive registration algorithm may result in poor outcomes in MM-SSL scenarios. 

To address challenge (\textbf{\textit{i}}), we propose a modality all-in-one network based on vanilla U-Net \cite{unet}, which accepts data from all modalities to predict segmentation masks without the need for modality registration or extra network structures. This makes it applicable to MM scenarios. To tackle challenge (\textbf{\textit{ii}}), we design the Modality-Level Modulation Bank (MLMB) for modality-invariant feature learning and the Modality-Level Prototype Bank (MLPB) for modality-specific feature learning.
Previous works \cite{hyperfed, fedBN, PDF} have shown that feature-level modulation with customized mean and variance can mitigate feature shifts caused by non-iid data. Inspired by this, we assign learnable modulation parameters to each modality and store them in MLMB. MLMB gradually modulates the features of different modalities into a modality-invariant feature space, ensuring the network's feature representation remains consistent across diverse modalities. To capture modality-specific information, we introduce MLPB, where each modality has specific prototypes stored. The differentiation of these prototypes helps distinguish modality-specific information. By combining MLMB and MLPB, we enable comprehensive MM feature extraction and representation. When the network receives data from a specific modality, it retrieves the corresponding modulation layer from MLMB and the prototype from MLPB, integrating them as plug-ins for segmentation prediction, as shown in Fig.~\ref{G_Net}. To update the MLMB and MLPB, we propose the \textbf{\textit{M}}odality \textbf{\textit{P}}rototype \textbf{\textit{C}}ontrastive \textbf{\textit{L}}earning (\textbf{\textit{MPCL}}) strategy.

Furthermore, different modalities exhibit varying learning efficiency, with some being easier to learn than others, which affects convergence speed. To address this, we design \textbf{\textit{M}}odality-\textbf{\textit{A}}daptive-\textbf{\text{W}}eighting (\textbf{\textit{MAW}}) to dynamically adjust learning weights. MAW balances learning speeds by assigning higher weights to lower-performing modalities and lower weights to better-performing ones.
To address challenge (\textbf{\textit{iii}}), we design a \textbf{\textit{D}}ual \textbf{\textit{C}}onsistency (\textbf{\textit{DC}}) SSL training strategy from both image and feature perspectives, without the reliance on generative methods. 
Image-perturbation consistency adds disturbances (e.g., noise) to the input images and ensures consistency between predictions from perturbed and unperturbed images. Feature-perturbation consistency randomly drops neurons in the encoder, feeds the perturbed intermediate features into the decoder, and enforces consistency between the predictions of perturbed and unperturbed features.

With these components, we construct a complete MM-SSL training framework, dubbed \textbf{\textit{D}}ouble \textbf{\textit{B}}ank \textbf{\textit{D}}ual \textbf{\textit{C}}onsistency (\textbf{\textit{DBDC}}).
The main contributions of this are summarized as: (\textbf{\textit{1}}) We propose an all-in-one network for MM-SSL medical image segmentation that supports any number of modalities. To the best of our knowledge, this is the first attempt to address scenarios with more than two modalities.
(\textbf{\textit{2}}) We introduce two plug-in banks for modality-invariant and modality-specific feature learning with modality prototype contrastive learning, enabling a comprehensive MM feature space.
(\textbf{\textit{3}}) We design a modality-adaptive weighting mechanism to balance the learning speeds across modalities, resulting in a stable MM learning process.
(\textbf{\textit{4}}) We introduce a dual-consistency learning strategy to make better use of unlabeled data for more reliable SSL training.

\section{Related Works}
\subsection{SSL-based Medical Image Segmentation}
Consistency-based training paradigms have achieved great success in SSL medical image segmentation by enforcing prediction consistency under various perturbations, such as input, feature, and network perturbations. Tarvainen \textit{et al.}~\cite{MeanTeacher} proposed Mean Teacher (MT), a representative consistency-based strategy using input perturbation. MT assumes that networks should produce consistent predictions for similar data, enforcing consistency on predictions from input data perturbed by random noise. To improve training stability, MT consists of two networks: a student model, updated via feedback during training, and a teacher model, whose weights are updated using an exponential moving average \cite{EMA} (EMA) of the student model's weights.
Unlike MT, Cross Pseudo Supervision~\cite{CPS} (CPS) employs a network perturbation strategy. It imposes prediction consistency between two segmentation networks initialized differently for the same input image. For labeled data, the outputs of both networks are supervised by their respective ground-truth segmentation masks. For unlabeled data, these outputs are supervised using pseudo-segmentation masks generated by the opposing network. To enhance the reliability of pseudo-label, Himashi \textit{et al.} \cite{peiris2023uncertainty} introduced uncertainty estimation into SSL.  This approach filters out high-uncertainty labels that may mislead the model, thereby improving the quality of the pseudo-labels and enhancing the stability of consistency learning. 

\subsection{Cross-Modality Medical Image Segmentation}
Cross-modality segmentation in medical image analysis is challenging due to distribution differences between modalities. To address this problem, Valindria \textit{et al.}~\cite{valindria2018multi} proposed a dual-stream architecture for cross-modality segmentation. This architecture has an independent encoder and decoder for each modality, with a shared bottleneck layer connecting them to enable cross-modality information fusion. This approach inspired a series of works \cite{multi1, multi2} that combine modality-specific and shared layers.
Dou \textit{et al.} \cite{douqi} proposed an unpaired MM segmentation method for CT and MRI images, using a new loss function based on knowledge distillation to improve cross-modality information fusion.
Disentangling representation is another way to address this challenge. Pei \textit{et al.}~\cite{pei2021disentangle} proposed a transfer learning method that disentangles each modality into modality-invariant and modality-specific features using two encoders. The modality-invariant features are shared and used for segmentation. Guo \textit{et al.}~\cite{causal} introduced causal learning to cross-modality medical image segmentation. They posited that anatomical structures represent causal knowledge, invariant across modalities, while modality factors (e.g., contrast) are non-causal and cause modality confounding. Similarly, they disentangled images into anatomical features and modality factors, designing a new loss function based on causal learning to improve disentangling representations.

\subsection{Prototype Learning}
Prototype learning originates from clustering algorithm~\cite{cover1967nearest} (e.g., nearest neighbors) in machine learning and prototype theory~\cite{knowlton1993learning, rosch1973natural} in cognitive science. Prototype learning groups features into clusters and uses the cluster centers as prototypes. The relevance of data to each category is determined by measuring the distance between features and prototypes. A smaller distance indicates higher relevance to a category, and vice versa. In image segmentation, several prototype learning-based methods have been proposed. Li \textit{et al.}~\cite{li2021adaptive} considered that similar and adjacent pixels can be represented by superpixels, which serve as prototypes for different categories. During the inference phase, each pixel is assigned to a superpixel, which determines its corresponding category. Zhou \textit{et al.}~\cite{zhou2022rethinking} rethought previous prototype learning methods and proposed a non-learnable prototype scheme for image segmentation. Specifically, they used an online clustering approach, which allows pixels to be clustered and their corresponding prototypes updated in each iteration. 

\begin{figure*}[]
    \centering
    \includegraphics[width=\textwidth]{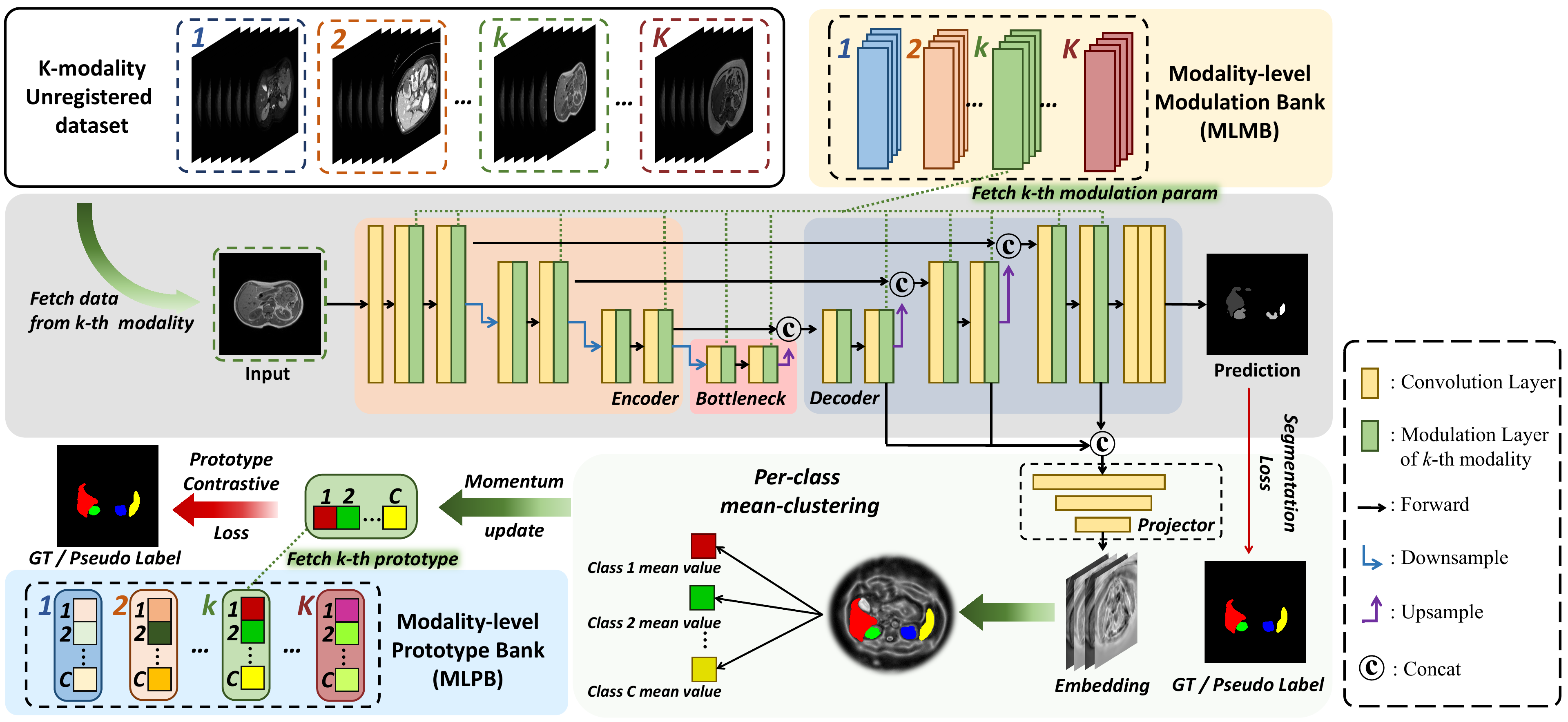}
    \caption{The structure of our proposed modality-all-in-one network, which is built on a vanilla U-Net, plug-in with the Modality-Level Modulation Bank (MLMB) and Modality-Level Prototype Bank (MLPB).}
    \vspace{-15pt}
    \label{G_Net}
\end{figure*}

\section{Methods}
Given a collection of unregistered datasets with $K$ different modalities, denoted as ${D_1, \dots, D_K}$. Each dataset consists of labeled and unlabeled parts, represented as $D_k = \{D_k^l, D_k^u\}$, where $D_k^l = \{(x_k^l, y_k^l)\}^{M_k^l}_{i=1}$ and $D_k^u = \{x_k^u\}^{M_k^u}_{i=1}$. Here, ${M_k^l}$ and ${M_k^u}$ denote the numbers of labeled and unlabeled samples for the $k$-th modality. In SSL tasks, it is common that $M_k^u \gg M_k^l$. 

\subsection{Modality All-in-One Network}
As shown in Fig. \ref{G_Net}, the modality all-in-one network $\mathcal{M}$ is built on the standard U-Net \cite{unet} architecture, with all convolution layers shared across modalities.
MLMB and MLPB store modality-level modulation and prototype parameters, which are used in the network as plug-ins. To be specific, the network selects the corresponding modulation and prototype parameters for each modality, facilitating the extraction of both modality-invariant and modality-specific features. We propose \textbf{\textit{M}}odality \textbf{\textit{P}}rototype \textbf{\textit{C}}ontrastive \textbf{\textit{L}}earning (\textbf{\textit{MPCL}}) to update MLMB and MLPB. Additionally, due to varying learning difficulties across modalities, we design \textbf{\textit{M}}odality-\textbf{\textit{A}}daptive \textbf{\textit{W}}eighting (\textbf{\textit{MAW}}) to dynamically adjust the learning weights for each modality. These components are explained in detail later.

\subsubsection{Modality-Level Modulation Bank (MLMB)}
Modality-specific means and variances for each modulation layer are learned and stored in MLMB, which are then used to normalize the feature of the corresponding modality. By applying multiple layers of modulation, feature shifts across modalities are gradually mitigated, enabling the extraction of modality-invariant features. 
The process for the $i$-th modulation layer of MLMB can be represented as:

\begin{equation}
    \text{MLMB}_i(z_i,k) = \gamma_{i,k} \cdot \frac{z_i-\mu_{i,k}}{\sqrt{\sigma_{i,k}^2+\epsilon}} + \beta_{i,k},
\end{equation}
where $z_i$ represents the features of the $i$-th convolution layer, $\gamma_{i,k}$ and $\beta_{i,k}$ are the affine parameters, and $(\mu_{i,k}, \sigma_{i,k}^2)$ represent the mean and variance for modality $k$. The term $\epsilon > 0$ is a small constant added to prevent numerical instability. The means $\mu_{i,k}$ and variances $\sigma_{i,k}^2$ are initially set to 0 and 1 respectively, and are updated using the Exponential Moving Average (EMA) method~\cite{EMA, zhou2022generalizable} during training:
\begin{align}
    \Bar{\mu}_{i,k}^{t+1} &= (1 - \alpha) \cdot \Bar{\mu}_{i,k}^t + \mu_{i,k}^t, \\
    (\Bar{\sigma}_{i,k}^{t+1})^2 &= (1 - \alpha) \cdot (\Bar{\sigma}_{i,k}^t)^2 + (\sigma_{i,k}^t)^2,
\end{align}
where $\alpha$ is the momentum coefficient and is empirically set to 0.99 in the experiments.

\subsubsection{Modality-Level Prototype Bank (MLPB)}
As shown in Fig. \ref{G_Net}, we concatenate each layer of the decoder to obtain a multi-scale feature, which is then passed through a 3-layer 1$\times$1 convolution projector to produce a $D$-dimension pixel embedding $I_k\in\mathbb{R}^{D \times N}$ for the $k$-th modality. Here, $I_k={[i^k_n]}^N_{n=1}$, $N=H\times W$, and $i^k_n \in\mathbb{R}^{D}$. We set $D=64$ in our experiments. We assign $K$ different prototypes $\{P_{k}\}_{k=1}^{K}$ for the $K$ modalities. By diversifying the prototypes, we also diversify the modality-specific information. Specifically, the prototype of each modality can be divided into multiple subclusters, with each cluster corresponding to a segmentation class. For a $C$-class segmentation task, it can be represented as $P_k = [p_c^k]^C_{c=1}$.

\paragraph{Prototype Initialization and Updating}
Given pixel embeddings $I_k={[i^k_n]}^N_{n=1}$ generated from data $x_k$ of the $k$-th modality, our goal is to map these embeddings to the corresponding prototype $P_k=\{p^k_{c}\}^C_{c=1}$. This embedding-to-prototype mapping is denoted as $L_{k}={[l_n]}^N_{n=1} \in \{0,1\}^{C\times N}$, where $l_n$ is the one-hot assignment vector for pixel $n$. The optimization of $L_{k}$ is performed by maximizing the similarity
between pixel embeddings $I_k$ and the prototype of $k$-th modality $P_k={[p^k_{c}]}^C_{c=1}$, expressed as:
\begin{equation}
\begin{aligned}
    \max_{L_k} & \text{Tr}(L_k^\top P_k^\top I_k),
    \\
    s.t. L_k \in \{0,1\}^{C\times N}, & L_k^\top\textbf{1}^C=\textbf{1}^N, L_k\textbf{1}^N=\frac{N}{C}\textbf{1}^C,
\end{aligned}
\end{equation}
where $\textbf{1}^C$ denotes the $C$-dimension ones vector.
The unique assignment constraint $L_k^\top\textbf{1}^C=\textbf{1}^N$ ensures that each pixel embedding is assigned to exactly one prototype. The equipartition constraint $L_k\textbf{1}^N=\frac{N}{C}\textbf{1}^C$ enforces that each prototype is selected approximately $\frac{N}{C}$ times on average in each forward operation. To solve Eq. (4), $L_k$ can be relaxed to be an element of the transportation polytope \cite{polytope}, as follows:
\begin{equation}
\begin{aligned}
    &\max_{L_k} \text{Tr}(L_k^\top P_k^\top I_k) + \varphi h(L_k),
    \\
    s.t. L_k \in & \mathbb{R}^{C\times N}_+, L_k^\top\textbf{1}^C=\textbf{1}^N, L_k\textbf{1}^N=\frac{N}{C}\textbf{1}^C,
\end{aligned}
\end{equation}
where $h(L_k)=\sum_{n,c}-l_{i_{n,c}}\log l_{i_{n,c}}$ and $\varphi$ is a parameter to control the smoothness. The solver can then be written as:

\begin{equation}
L_k=\text{diag}(u)\text{exp}(\frac{P_k^\top I_k)}{\varphi})\text{diag}(v), 
\end{equation}
where $u \in \mathbb{R}^C$ and $v \in \mathbb{R}^N$ are renormalization vectors,
computed using Sinkhorn-Knopp iteration \cite{sinkhorn}. The prototypes are continuously updated using an online scheme with EMA after each training iteration, as follows:
\begin{equation}
    p_c^k = \alpha \cdot p_c^k + (1-\alpha)\cdot\Bar{i}_c^k,
\end{equation}
where $\alpha$ is the momentum coefficient, empirically set to 0.99 in the experiments, and $\Bar{i}_c^k$ is the mean vector of the pixel embeddings.

\begin{figure*}[]
    \centering
    \includegraphics[width=0.9\textwidth]{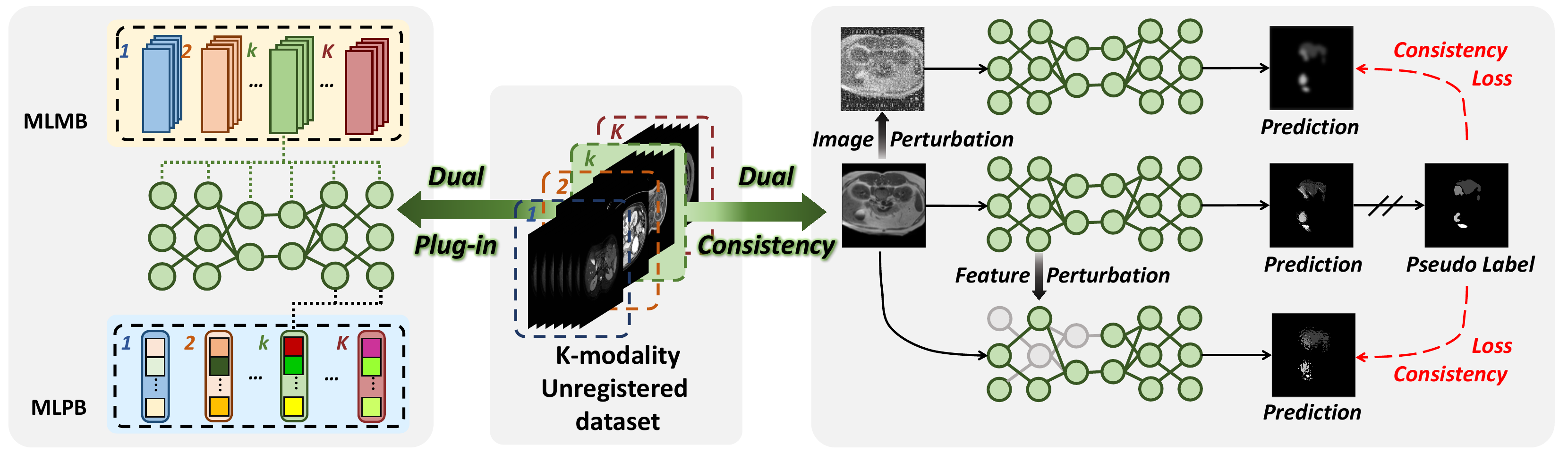}
    \caption{Pipeline of our proposed Double Bank Dual Consistency (DBDC) framework.}
    \label{framework}
\end{figure*}

\paragraph{Modality Prototype Contrastive Learning (MPCL)}
The cosine similarity $s_k$ between the pixel embedding and the prototype for the $k$-th modality can be calculated as $s_k = \langle I_k, P_k \rangle$, where $\langle\cdot\rangle$ represents cosine similarity.
Then, we use a combination of dice loss and cross-entropy loss to calculate the prototype segmentation loss:
\begin{equation}
    \mathcal{L}_{pseg} = \frac{1}{2}\mathcal{L}_{dice}(s_k, y^l_k) + \frac{1}{2}\mathcal{L}_{ce}(s_k, y^l_k).
\end{equation}

To differentiate the prototypes across modalities, we design a modality contrastive loss, represented as $\mathcal{L}_{mc}$. This loss maximizes the cosine similarity between the generated embedding and its corresponding prototype while minimizing the similarity with other prototypes. It is defined as:
\begin{equation}
    \mathcal{L}_{mc} = -\langle I_k, P_k \rangle + \langle I_k, P_{j\ne k} \rangle.
\end{equation}

Note that maximizing cosine similarity is equivalent to minimizing negative cosine similarity, and vice versa.

To optimize inter-class relationships, we apply prototype contrastive loss $\mathcal{L}_{pc}$ to encourage the pixel embedding $i_k$ to be similar to its assigned-class prototype $p^k_c$ (the positive prototype sample) and dissimilar to the remaining $C-1$ prototypes, viewed as negative prototype samples $\mathcal{P}^-$. The prototype contrastive loss $\mathcal{L}_{pc}$ is defined as:
\begin{equation}
    \mathcal{L}_{pc} = - \log\frac{\exp({i_k^\top p^k_c}/\tau)}{\exp({i_k^\top p_c^k}/\tau)+\sum_{p^-\in\mathcal{P}^-}\exp({i_k^\top p^-}/\tau)}.
\end{equation}

To optimize intra-class relationships, we use prototype distance loss $\mathcal{L}_{pd}$ to compact pixel features belonging to the same prototype. It is defined as:
\begin{align}
    \mathcal{L}_{pd}={(1-i_k^\top p_c^k)}^2.
\end{align}

The total loss for MPCL is formulated as:
\begin{equation}
    \mathcal{L}_{MPCL}=(\mathcal{L}_{pseg}+\mathcal{L}_{mc}+\mathcal{L}_{pc}+\mathcal{L}_{pd})/4.
\end{equation}

\subsubsection{Modality-Adaptive Weighting (MAW)}
Due to inherent disparities in learning complexity across data modalities, neural networks often exhibit varying convergence speeds across these modalities, resulting in imbalanced learning. To address this issue, we introduce the Modality Adaptive Weighting (MAW) strategy to balance learning across modalities. MAW evaluates modality imbalance from two aspects: learning speed and training performance. Specifically, the weighting of each modality is inversely proportional to its learning rate and directly proportional to its error rate. For the $k$-th modality at the $t$-th epoch, the weight is defined as:
\begin{equation}
    \omega_{k,t} = \frac{1}{\mathcal{S}_{k,t}^\beta} \cdot (1-\mathcal{P}_{k,t}), 
\end{equation}
where $\beta$ is empirically set to 0.2 in the experiments to alleviate outliers. We use the Population Stability Index \cite{PSI} to measure the learning speed $\mathcal{S}_k$ for $k$-th modality at the $t$-th epoch, based on the ratio of positive to negative training times over a fixed number of epochs, defined as:
\begin{align}
    \mathcal{S}_{k,t} &= \frac{d^+_{k,t} + \epsilon}{d^-_{k,t} + \epsilon},
    \\[-0.3em]
    d^+_{k,t} &= \sum_{t-\tau}^t \mathds{I} (\bigtriangleup > 0) ln (\frac{\lambda_{k,t}}{\lambda_{k,t-1}}),
    \\[-0.3em]
    d^-_{k,t} &= \sum_{t-\tau}^t \mathds{I} (\bigtriangleup \le 0) ln (\frac{\lambda_{k,t}}{\lambda_{k,t-1}}), 
\end{align}
where $\epsilon$ is a smoothing item with minimal value, empirically set to 1e-8, and $\lambda_{k,t}$ denotes the $DSC$ score of the $k$-th modality at the $t$-th epoch. $\bigtriangleup=\lambda_{k,t} - \lambda_{k,t-1}$ represent the change in the $DSC$ between adjacent epochs for the $k$-th modality. $\mathds{I}$ is the indicator function that separates the positive and negative training for the $k$-th modality.
$\tau$ is the number of accumulation epochs, empirically set to 20.
Modalities that learn faster at the $t$-th epoch have larger $\mathcal{S}_{k,t}$, resulting in smaller corresponding weights in the loss function, and vice versa.
To measure training performance, we ccumulate $\lambda_{k,t}$ over $\tau$ epochs to obtain $\mathcal{P}_{k,t}$. The MAW for all modalities is then defined as $W=[\omega_1, \dots, \omega_K]$.

\begin{table*}[]
\centering
\caption{Comparison with different cross-modality methods on 2-modality heart segmentation dataset.}
\begin{tabular}{c|cc|cc|cc|c|c}
\hline\hline
 & \multicolumn{2}{c|}{CT} & \multicolumn{2}{c|}{MR} & \multicolumn{2}{c|}{Mean} & \multirow{2}{*}{Parms} & \multirow{2}{*}{FLOPs}\\
 & $DSC$ & $HD_{95}$ & $DSC$ & $HD_{95}$ & $DSC$ & $HD_{95}$  &&\\ \hline
 UNet (LB)&69.25&10.69 &59.34&18.03 &64.30&14.36 &4.88 &5.67\\
 UNet (UB)&89.46&2.78 &85.40&4.83 &87.43&3.81 &4.88 &5.67\\ \hline
 VarDA &51.77&17.62 &29.18&36.18 &33.47 &30.39 &33.28 &38.7\\
 VarDA$\star$ &63.32&9.08 &53.78&15.05 &58.55&12.06 &33.28 &38.7\\
 RAM-DSIR &46.08&16.25 &13.85&41.78 &29.96 &29.01 &3.80 &1.84\\
 RAM-DSIR$\star$ &59.55&9.82 &31.31&36.68 &45.43&23.25 &3.80 &1.84\\
 SUMML &84.28&4.37 &66.05 &8.18 &75.17&6.28 &25.02&23.46\\
 MASS &79.09&5.49 &70.43&6.86 &74.76&6.18 &0.72 &61.44\\ \hline
 \rowcolor{lightergray}
 Ours w/o MLPB$\And$MAW &86.73 &3.33 &78.65&5.80 &82.69&4.57 &4.91&6.05\\
  \rowcolor{lightergray}
 Ours w/o MAW  & \textbf{87.40} & \textbf{3.03}&\underline{80.54}&\underline{5.35}&\underline{83.97}&\underline{4.19} &5.34&20.03     \\ 
  \rowcolor{lightergray}
  Ours & \underline{87.37}& \underline{3.30}&\textbf{81.74}&\textbf{5.11}&\textbf{84.56}&\textbf{4.21} &5.34&20.03\\
 \hline\hline
\end{tabular}
\label{MMWHS_tab}
\end{table*}

\subsection{Dual Consistency Training Stategy}
We enforce both image and feature perturbation consistency on unlabeled data. The dual consistency loss is defined as:
\begin{align}
    \mathcal{L}_{dc}^k &= \mathcal{L}_{ipc}^k + \mathcal{L}_{fpc}^k,
    \\
    \mathcal{L}_{ipc}^k = \mathcal{L}_{ce}(\mathcal{M}(\hat{x}_k^u;k), Y) &+ \delta \mathcal{L}_{MPCL}(\mathcal{M}(\hat{x}_k^u;k), Y),
    \\
    \mathcal{L}_{fpc}^k = \mathcal{L}_{ce}(\mathcal{\Check{M}}({x}_k^u;k), Y) &+ \delta \mathcal{L}_{MPCL}(\mathcal{\Check{M}}(\hat{x}_k^u;k), Y),
\end{align}
where $\mathcal{L}_{ipc}^k$ and $\mathcal{L}_{fpc}^k$ represent the image and feature perturbation consistency losses for the $k$-th modality, respectively. $\hat{x}_k^u$ is the image-perturbated unlabeled data and $Y$ is the pseudo label generated by $\mathcal{M}({x}_k^u;k)$. $\Check{\mathcal{M}}$ is the perturbated network, and $\Check{\mathcal{M}}({x}_k^u;k)$ represents the feature perturbation process. $\delta$ is a balancing parameter, empirically set to 0.1 in the experiments.

As shown in Fig. \ref{framework}, for the $k$-th modality, the combined supervision and dual consistency loss, considering both labeled data $x_k^l$ and unlabeled data $x_k^u$, is written as:
\begin{equation}
    \mathcal{L}_{modal}^k = \mathcal{L}_{sup}(x_k^l, y_k^l) + \lambda \mathcal{L}_{dc}(x_k^u),
\end{equation}
where $\mathcal{L}_{sup}$ is the supervision loss, and $\mathcal{L}_{dc}$ is the dual consistency loss. $\lambda$ is a coefficient that balances $\mathcal{L}_{sup}$ and $\mathcal{L}_{con}$, updated using a Gaussian ramp-up function $\lambda (t) = \lambda_{max} * e^{-5 \left(1.0 - \frac{t}{t_{max}}\right)^{2}}$, where $\lambda_{max} = 1.0$ and $t_{max}$ is the maximum epoch number. The supervision loss for the $k$-th modality is written as:
\begin{equation}
\begin{aligned}
    \mathcal{L}_{sup}^k = \mathcal{L}_{ce}(\mathcal{M}(x_k^l;k),y_k^l) + \mathcal{L}_{dice}(\mathcal{M}(x_k^l;k),y_k^l) \\
    + \delta \mathcal{L}_{MPCL}(x_k^l;k),y_k^l),
\end{aligned}
\end{equation}
where $\mathcal{L}_{ce}$ is the cross-entropy loss, $\mathcal{L}_{dice}$ is the dice loss, and $\delta$ is a balancing parameter set to 0.1 in the experiments. Therefore, the total loss of our framework is:
    \begin{align}
\mathcal{L}_{total}&=\sum_{k=1}^K\omega_k\mathcal{L}^k_{modal} \\
&=\sum_{k=1}^K \omega_k(\mathcal{L}_{sup}(x_k^l, y_k^l)+\lambda \mathcal{L}_{con}(x_k^u)).
    \end{align}

In this way, our method leverages unlabeled data to enhance the generalization and employs MLMB and MLPB for MM learning.

\begin{table*}[]
\centering
\caption{Comparison with different single-modality SSL methods at 20\% labeled ratio on the 4-modality abdominal dataset. `ip' and `op' represent `in-phase' and `out-phase', respectively. `LB' represents the lower bound, which is trained on 20\% labeled data, and `UB' represents the upper bound, which is trained on fully labeled data, respectively.}
\centering
\begin{tabular}{c|cc|cc|cc|cc|cc|c|c}
\hline\hline
 &\multicolumn{2}{c|}{T2} & \multicolumn{2}{c|}{T1 ip} & \multicolumn{2}{c|}{T1 op} & \multicolumn{2}{c|}{CT} & \multicolumn{2}{c|}{Mean} & \multirow{2}{*}{Parms} & \multirow{2}{*}{FLOPs}\\
 &$DSC$&$HD_{95}$ &$DSC$&$HD_{95}$ &$DSC$&$HD_{95}$ &$DSC$&$HD_{95}$ &$DSC$&$HD_{95}$ &&\\ \hline
UNet (LB) &82.80&12.21 &68.51&23.91&69.71&14.57&76.52&17.72&74.39&17.10 &4.88 &10.07\\
UNet (UB) &94.77&1.06&92.97&1.39&92.37&1.15&88.47&8.43&92.14&3.01 &4.88 &10.07\\ \hline
 MT &84.40&12.41&66.93&20.03&69.23&21.47&66.83&26.22 & 71.85&20.03 &9.75 &20.14\\ 
 UAMT &88.60&11.44 &68.68&15.08 &68.85&22.57 &69.02&17.76 &73.78&16.71 &9.75 &20.14\\
CPS&88.52&10.28&77.12&6.64&71.50&13.71&82.32&17.01&79.87&11.91 &9.75 &20.14\\
 ICT&86.02&15.02 &74.62&5.50 &76.50&15.31 &83.12&15.26 &80.06&12.77 &9.75 &20.14\\
 EVIL&90.12&10.18 &78.72&12.72 &74.73 &18.11 &82.45 &15.20 &81.50&14.06 &9.75 &20.14\\\hline
  \rowcolor{lightergray}
 Ours w/o MLPB$\And$MAW &\textbf{92.14}&\textbf{3.97} &83.71&3.60 &79.53 &9.57 &82.79 &15.37 &84.54 &8.13 &4.91 &10.74\\
 \rowcolor{lightergray}
 Ours w/o MAW&89.96&9.85& \underline{85.18} & \underline{3.23}& \underline{84.61} & \underline{2.92} &\textbf{86.91}&\underline{14.61}&\underline{86.67} &\underline{7.65} &5.34 &35.58\\ 
  \rowcolor{lightergray}
 Ours & \underline{92.10} & \underline{4.34}& \textbf{86.85} &\textbf{2.91}& \textbf{85.63} & \textbf{2.60} &\underline{85.86}& \textbf{8.61} & \textbf{87.61} & \textbf{4.62} &5.34 &35.58\\
 \hline\hline
\end{tabular}
\label{CHAOS_tab}
\end{table*}

\section{Experiment}
\subsection{Dataset}
We create two MM medical image segmentation datasets based on three publicly available benchmark datasets.
\subsubsection{Heart dataset}: We construct a 2-modality heart dataset from
the \textbf{\textit{Multi-Modality Whole Heart Segmentation (MM-WHS)}} dataset \cite{MMWHS}, which includes CT and MR modality, each containing 20 3D scans. The scans cover the entire heart substructures: left and right ventricle blood cavity, left and right atrium blood cavity, myocardium of the left ventricle, ascending aorta, and pulmonary artery. Following \cite{MMWHS}, we crop the region of interest (ROI) for each scan and resample all 3D scans into 192 $\times$ 192 2D images. Each modality is randomly split into 70\% for training, 10\% for validation, and 20\% for testing. For the SSL setting, we randomly sample 20\% of the training data.

\subsubsection{Abdominal dataset}: We construct a 4-modality abdominal dataset, which includes CT, T2, T1 in-phase, and T1 out-phase modalities. The CT data are sourced from the \textbf{\textit{Multi-Atlas Labeling Beyond the Cranial Vault Challenge (BTCV)}} dataset \cite{BTCV}, which contains 50 3D scans. The T2, T1 in-phase, and T1 out-phase data are sourced from the \textbf{\textit{Combined Healthy Abdominal Organ Segmentation (CHAOS)}} data \cite{CHAOS}, with each containing 40 3D scans. The combined abdominal dataset includes four abdominal organs: liver, left kidney, right kidney, and spleen. All 3D scans are resampled into 256 $\times$ 256 2D images.
Each modality is randomly split into 70\% for training, 10\% for validation, and 20\% for testing. For the SSL setting, we randomly sample 20\% of the training data.

\begin{figure*}[!t]
    \centering
    \includegraphics[width=\textwidth]{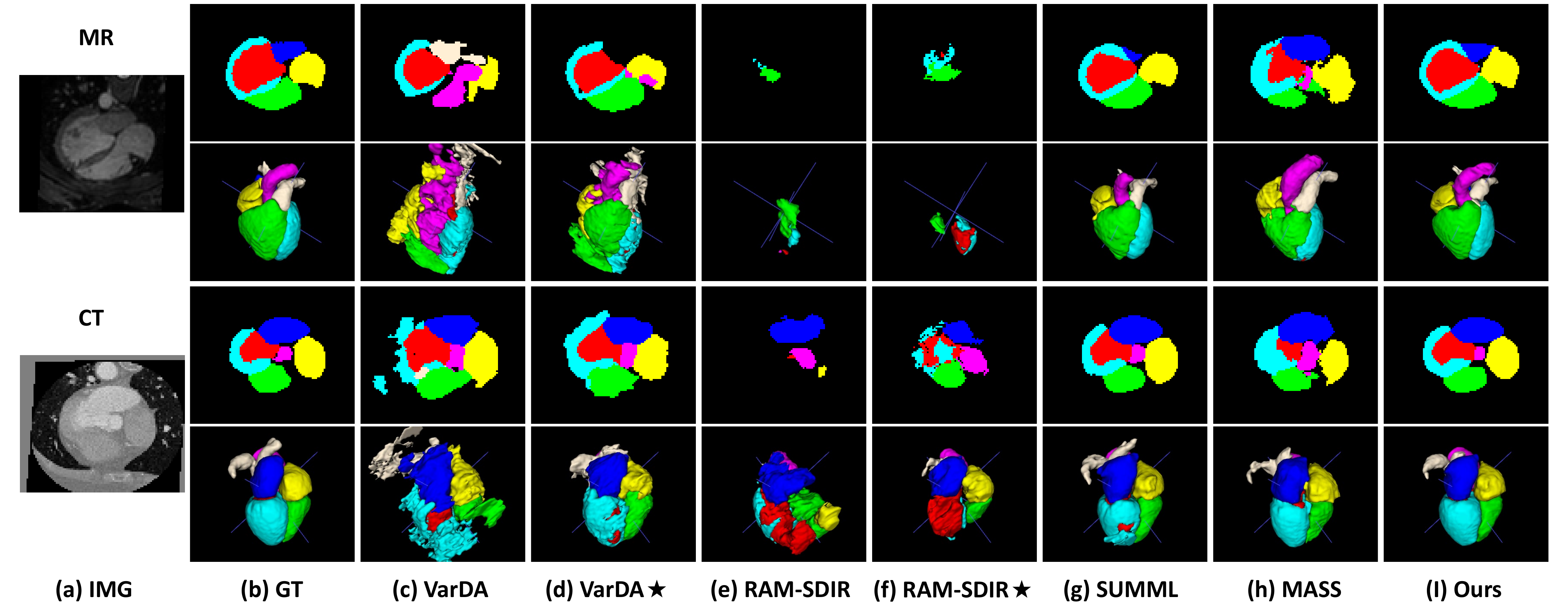}
    \caption{2D and 3D segmentation results on the 2-modality dataset.}
    \label{MMWHS_fig}
\end{figure*}

\begin{figure*}[!t]
    \centering
    \includegraphics[width=\textwidth]{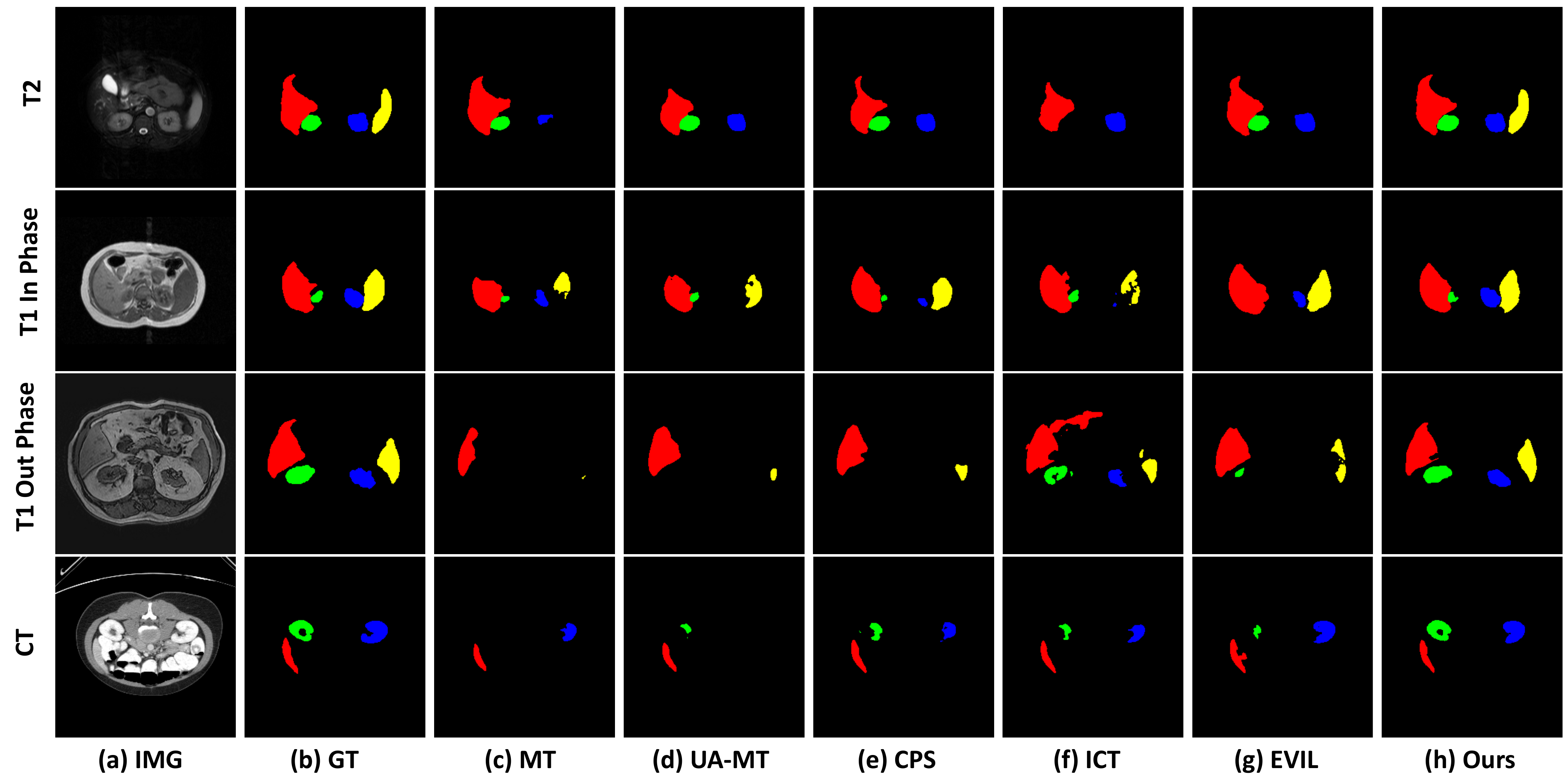}
    \caption{Segmentation results on the abdominal dataset.}
    \label{CHAOS_fig}
\end{figure*}

\subsection{Experimental Setup}
\subsubsection{Baselines}
We select several state-of-the-art methods for comparison, including single-modality SSL methods such as MT \cite{MeanTeacher}, UAMT \cite{UAMT}, CPS \cite{CPS}, ICT \cite{ICT}, and EVIL \cite{EVIL}; MM-SSL methods such as SUMML \cite{SUMML} and MASS \cite{MASS}; and transfer learning methods like VarDA \cite{VarDA} and RAM-SDIR \cite{RAM-DSIR}. All compared methods were implemented using their official codes and settings. 
\subsubsection{Training Details}
We set the epoch length to the maximum dataset length among the different modalities, conducting a total of 100 epochs.
Adam \cite{Adam} was utilized as the optimizer, with an initial learning rate of 1e-4 and a polynomial scheduler strategy. The proposed framework was implemented in PyTorch, using a single NVIDIA GTX 3090 GPU. The input images were randomly cropped to $128\times128$. A sliding window strategy was employed for validation and testing. Two popular metrics are used to evaluate performance: Dice Similarity Coefficient ($DSC$) and Hausdorff Distance ($HD_{95}$). $DSC$ quantifies overall accuracy, while $HD_{95}$ measures boundary precision. The combination of these two metrics offers a comprehensive evaluation of segmentation performance.

\begin{table}[!t]
\centering
\caption{The ablation study on the loss components.}
\label{ablation_proto_tab}
\begin{tabular}{c|c|c|c|c|cc}
\hline\hline
\multirow{2}{*}{$\mathcal{L}_{ipc}$} & \multirow{2}{*}{$\mathcal{L}_{fpc}$}
& \multirow{2}{*}{$\mathcal{L}_{mc}$} 
& \multirow{2}{*}{$\mathcal{L}_{pc}$} 
& \multirow{2}{*}{$\mathcal{L}_{pd}$} 
& \multicolumn{2}{c}{ours} \\  \cline{6-7}
&&&&& \multicolumn{1}{c|}{$DSC$} & $HD_{95}$ \\ \hline
$\times$&$\times$&$\times$&$\times$&$\times$ &\multicolumn{1}{c|}{81.88} &8.85  \\
$\surd$&$\times$&$\times$&$\times$&$\times$ &\multicolumn{1}{c|}{83.61} &7.39  \\ 
$\surd$&$\surd$&$\times$&$\times$&$\times$ &\multicolumn{1}{c|}{84.21} &7.80  \\ 
$\surd$&$\surd$&$\surd$&$\times$&$\times$ & \multicolumn{1}{c|}{85.25} &6.75  \\ 
$\surd$&$\surd$&$\surd$&$\surd$&$\times$ & \multicolumn{1}{c|}{\underline{86.02}} &\underline{5.61}  \\ 
$\surd$&$\surd$&$\surd$&$\surd$&$\surd$ & \multicolumn{1}{c|}{\textbf{87.61}} &\textbf{4.62}  \\ 
\hline\hline
\end{tabular}
\end{table}


\begin{figure*}[!t]
    \centering
    \includegraphics[width=.95\linewidth]{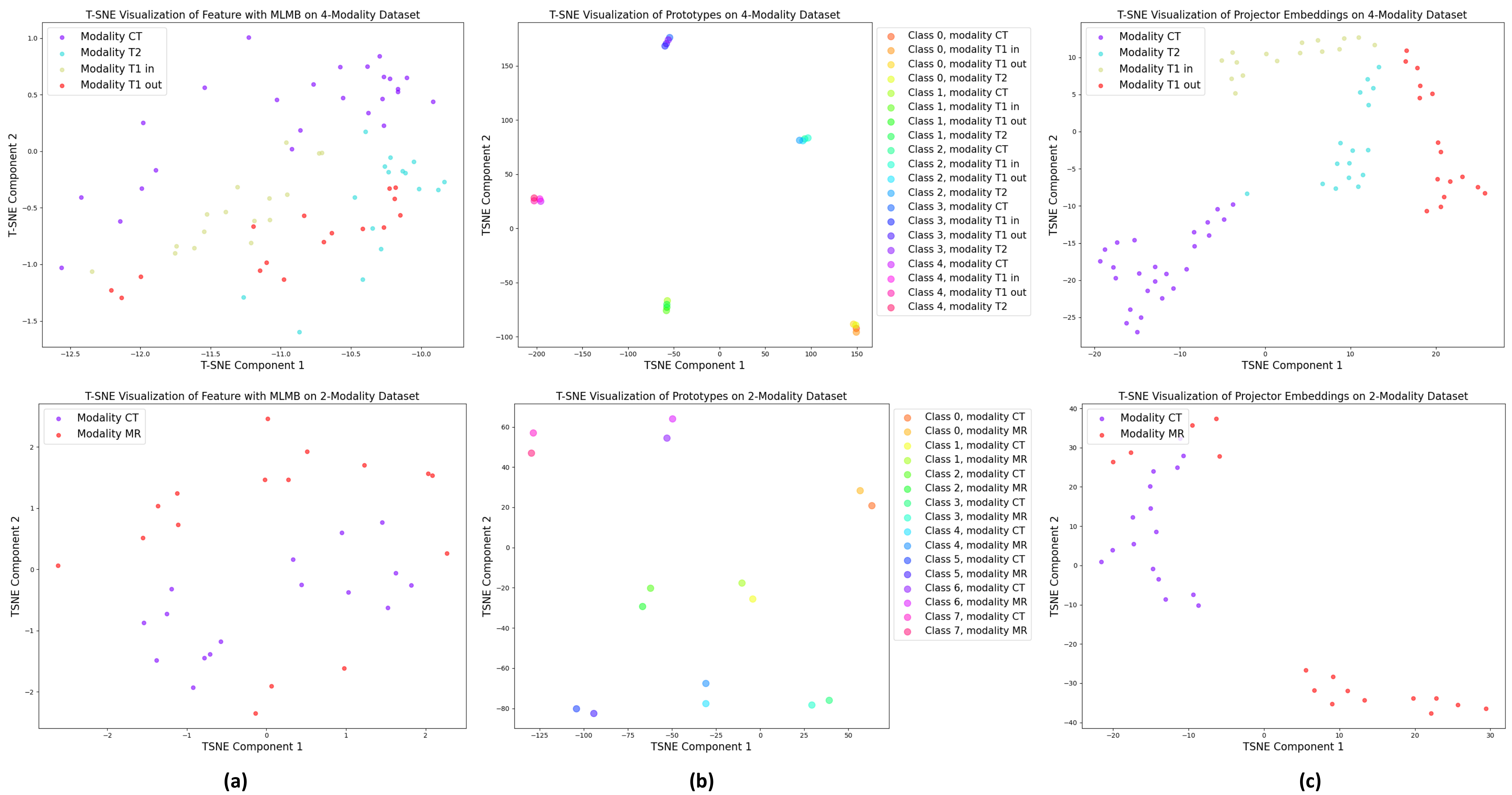}
    \caption{T-SNE visualization results, where the first and second rows represent the results on the 4-modality abdominal and 2-modality heart datasets, respectively. (a) represents the features after the last MLMB layers of the network. (b) represents the results of the MLPB. (c) represents the results of the embeddings after the projector.}
    \label{TSNE}
\end{figure*}

\begin{table}[!t]
\centering
\caption{The ablation study on the MLMB with different network layers.}
\label{ablation_msbn_tab}
\begin{tabular}{c|c|c|cc}
\hline\hline
\multirow{2}{*}{Encoder} 
& \multirow{2}{*}{Bottleneck} 
& \multirow{2}{*}{Decoder} 
& \multicolumn{2}{c}{ours} \\  \cline{4-5}
&&& \multicolumn{1}{c|}{$DSC$} & $HD_{95}$ \\ \hline
$\times$&$\times$&$\times$ &\multicolumn{1}{c|}{60.73} &33.95 \\ 
$\surd$&$\times$&$\times$ &\multicolumn{1}{c|}{83.52} &\underline{6.58} \\ 
$\surd$&$\surd$&$\times$ & \multicolumn{1}{c|}{\underline{84.93}} &6.91 \\ 
$\surd$&$\surd$&$\surd$ & \multicolumn{1}{c|}{\textbf{87.61}} &\textbf{4.62} \\ 
\hline\hline
\end{tabular}
\end{table}

\begin{figure}[!t]
    \centering
    \includegraphics[width=.9\linewidth]{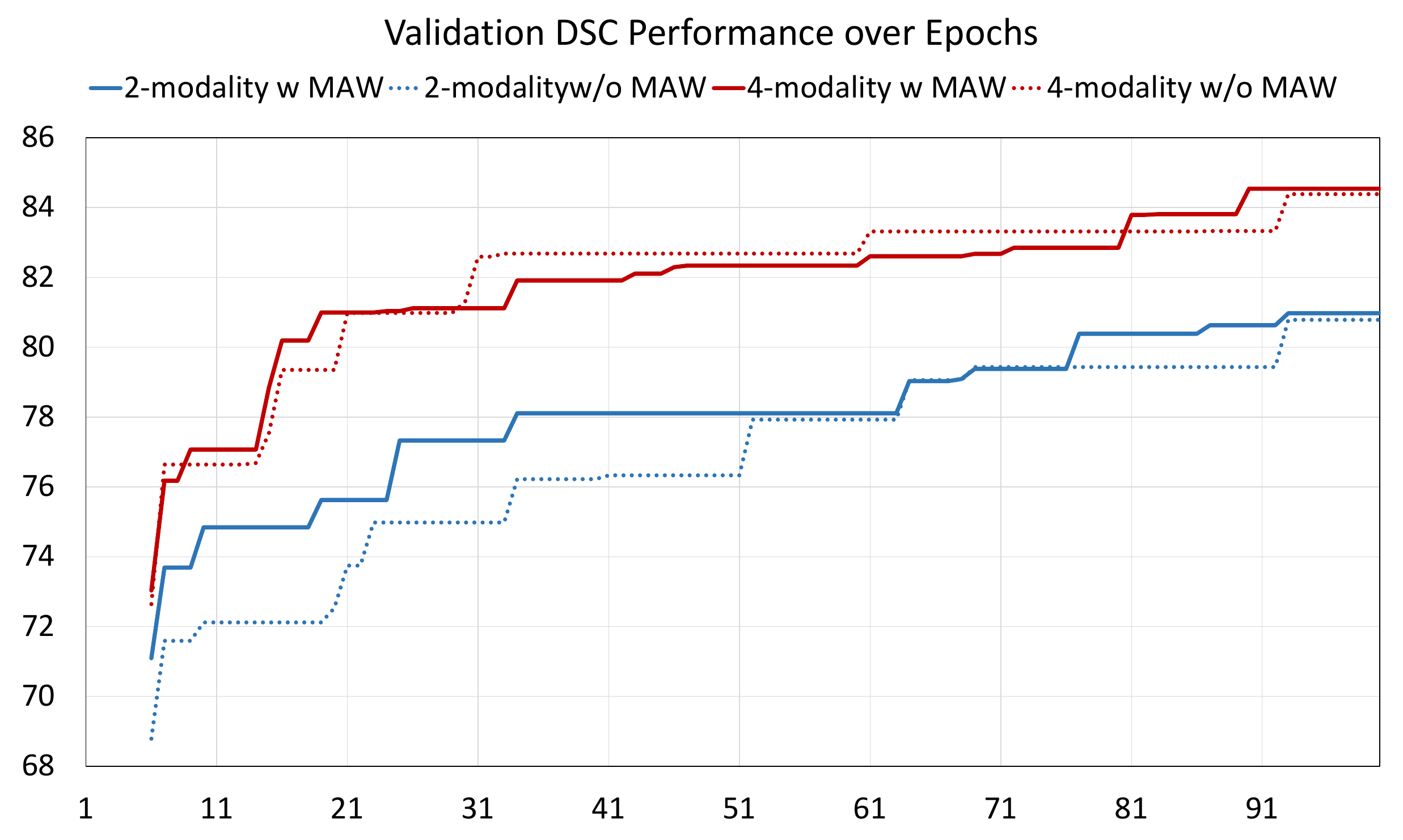}
    \caption{Mean DSC value on validation set over epochs.}
    \label{MADW}
\end{figure}

\begin{figure*}[!t]
    \centering
    \includegraphics[width=\linewidth]{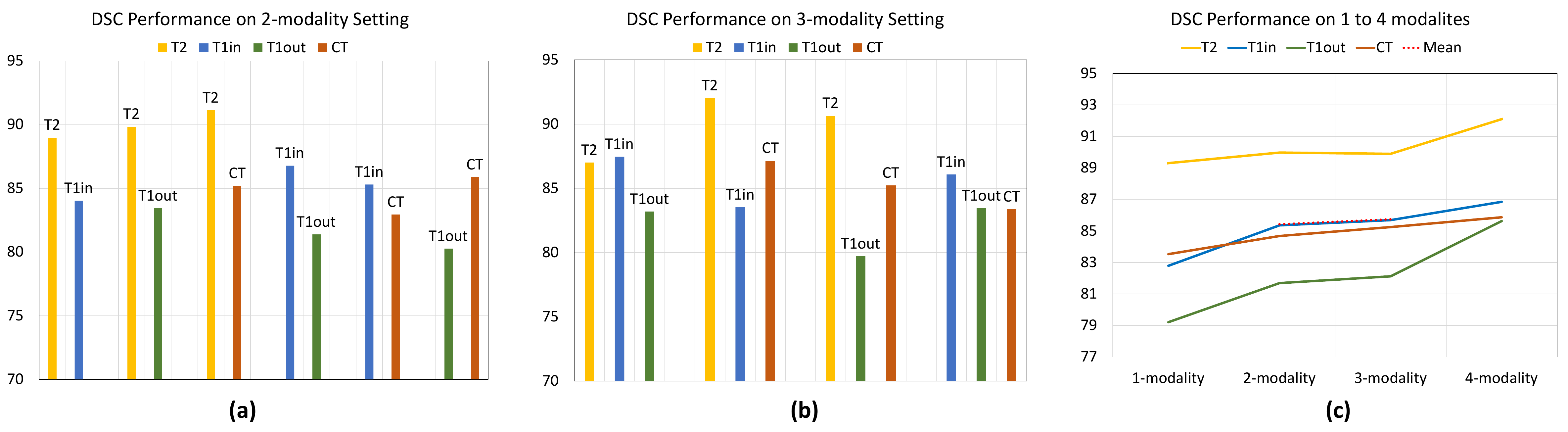}
        \caption{Results of modality number analysis on the abdominal dataset from 2 to 4 modalities. (a) Results of 2-modality combinations. (b) Results of 3-modality combinations. (c) DSC performance with adding modality numbers gradually.}
    \label{add_modalities}
\end{figure*}

\subsection{Main Results}

\subsubsection{Compared with unpaired MM baselines} Fig. \ref{MMWHS_fig} presents the segmentation results of different unpaired MM baselines on the heart dataset. It is obvious that our approach produces more accurate segmentation masks compared to other methods in both modalities. Transfer learning methods (VarDA, RAM-SDIR) struggle to accurately segment the target organs due to ineffective cross-modality learning, while MM semi-supervised methods (SUMML, MASS) exhibit numerous segmentation errors. Tab. \ref{MMWHS_tab} provides a quantitative comparison of the performance of various methods. We can observe that transfer learning methods (VarDA, RAM-DSIR) underperform compared to the backbone network's lower bound (LB), mainly due to the lack of labeled target modality data, regardless of whether the source modality data is labeled in a semi-supervised or fully supervised setting. The performance of transfer learning is severely hindered when there are large modality differences, such as between CT and MR, and further degrades with insufficient labeled source modality data.. MM-SSL methods perform better, but our approach still outperforms all other methods. The average $DSC$ of our method is nearly 10 points higher than that of SUMML and MASS, and 15 points higher in the MR modality compared to SUMML. Additionally, the introduction of MLPB and MAW gradually improves the $DSC$ and $HD_{95}$ metrics, significantly enhancing performance in the more challenging MR modality and balancing performance across modalities. Compared to the backbone's upper bound (UB), our method is only 2 points lower on CT and 3 points lower on MR.

\subsubsection{Compared with single-modality SSL baselines} Fig. \ref{CHAOS_fig} visualizes the segmentation results of our proposed method compared to other methods on the abdominal dataset. Our DBDC clearly achieves better segmentation results, more closely matching the ground truth, compared to all SSL medical image segmentation methods.
Tab. \ref{CHAOS_tab} reports the results of our approach trained on the abdominal dataset compared to other single-modality approaches trained separately on each modality. It is obvious that our approach consistently outperforms all single-modality methods, showing the advantages of MM learning, where different modalities compensate each other for the lack of labeled data in SSL.
Besides, we observe that when MLPB is not used for MM learning, the performance of T2 is significantly higher than the other modalities, with a $DSC$ score 10 points higher. This occurs because, at this stage, the all-in-one network focuses only on modality-invariant feature learning, favoring easier-to-learn modalities while neglecting harder-to-learn ones.
With the introduction of MLPB, the harder-to-learn modalities (e.g., T1 op) achieve an average $DSC$ improvement of over 5 points. This improvement results from the differentiation of modality-specific information enabled by MLPB, compensating for the learning capabilities in harder-to-learn modalities. 
However, the performance of T2 decreases with the introduction of MLPB. This is because MLPB enforces differentiation among modalities, which compromises the inherent advantages of the easier-to-learn modality. Therefore, we introduce MAW to balance learning difficulties across modalities. As shown in Tab. \ref{CHAOS_tab}, introducing MAW ensured the performance of the easier-to-learn modality, resulting in a 2-point DSC improvement compared to without MAW.

\subsection{Ablation Studies}

\subsubsection{Effectiveness of MLMB} Tab. \ref{ablation_msbn_tab} reports the results of gradually incorporating MLMB into the network, starting from the encoder and extending to the decoder. The table shows that the performance is poor when MLMB is not used, due to the network's inability to handle modality differences. However, when MLMB is applied only in the encoder, performance significantly improves, indicating that MLMB effectively mitigates modality distribution differences. Incorporating MLMB into both the bottleneck and decoder layers yields the best results. The modality distribution visualization is shown in Fig. \ref{TSNE} (a), where T-SNE is used to cluster the modality features from the last convolution layer with MLMB. The distances between samples of different modalities are very close, which means the feature distributions of different modalities become similar after several layers of MLMB. This demonstrates that MLMB effectively extracts modality-invariant features.

\subsubsection{Effectiveness of MMPCL}
Table \ref{ablation_proto_tab} presents the ablation experiment for MMPCL on the abdominal dataset. We incrementally add the proposed components to demonstrate their effectiveness.
It is evident that each component of MMPCL contributes positively. The introduction of modality contrastive loss $\mathcal{L}_{mc}$, prototype contrastive loss $\mathcal{L}_{pc}$, and prototype distance loss $\mathcal{L}_{pd}$ effectively strengthens both inter-modality and intra relationships, leading to significant performance improvements, with $DSC$ increasing by approximately 4 points).

\subsubsection{Effectiveness of MAW} First, as shown in Tab. \ref{MMWHS_tab} and \ref{CHAOS_tab}, the introduction of MAW successfully balanced the performance across different modalities. With MAW, the average $DSC$ and $HD_{95}$ scores improve by 1 point and 3 points, respectively. Second, from the perspective of learning speed, as shown in Fig. \ref{MADW}, validation performance improves more steadily and rapidly with MAW compared to without.

\subsubsection{Effectiveness of Dual-consistency strategy} As shown in the first, second and third rows of Tab. \ref{ablation_proto_tab}, the gradual introduction of image perturbation consistency loss $\mathcal{L}_{ipc}$ and feature perturbation consistency loss $\mathcal{L}_{fpc}$ results in progressively better outcomes, demonstrating the effectiveness of our dual-consistency training strategy.

\subsubsection{Modality Number Analysis}
To validate the benefits of learning with a larger number of modalities, we conducted a modality number analysis, as shown in Fig. \ref{add_modalities}. Fig. \ref{add_modalities}(a) and (b) show the results for all modality combinations when training with 2 or 3 modalities from a total of 4 available modalities on the abdominal dataset. Different modality combinations affect the performance of each modality in the current multimodal learning scenario. Fig. \ref{add_modalities}(c) shows that that as the number of modalities increases, the performance of each modality improves, with T2 and T1out showing a particularly significant improvement of over 3 points.

\subsubsection{Parameter Analysis} As shown in the last two columns of Tab. \ref{CHAOS_tab} and \ref{MMWHS_tab}, we compare the parameter quantities and FLOPs of different methods. 
SUMML is based on CycleGAN \cite{CycleGAN}, which introduces multiple networks, resulting in a significantly larger parameter quantity. MASS, by contrast, is implemented using a lightweight registration network \cite{warp}, resulting in fewer parameters. However, since MASS processes the entire 3D image, its FLOPs are relatively high. VarDA introduces multiple variational autoencoders \cite{VAE} to capture image features at different scales, which increases the parameter quantity as the number of networks grows.
Single-modality methods such as MT, UAMT, and CPS use two backbone networks (U-Net) for SSL consistency training, doubling the parameter quantity compared to a single backbone network. Our proposed method does not introduce additional networks. The extra parameters come solely from the addition of MLMB and MLPB.
As a result, our method remains lightweight, requiring only an additional 0.05M parameters for each added modality.

\section{Conclusion}
In this work, we propose a novel approach for MM-SSL medical image segmentation, with three key components: (1) A plug-in Double-Bank architecture designed for comprehensive MM learning; (2) A MAW mechanism for balanced and stable learning; and (3) A Dual-Consistency strategy to effectively leverage unlabeled data. 
Compared to previous work, our method significantly advances MM learning by supporting any number of modalities, which greatly enhances its generalization ability. 
This flexibility is enabled by the plug-in Double-Bank structure, allowing the network to process data from different modalities within a unified framework.
The Double-Bank architecture also facilitates the learning of both modality-invariant and modality-specific features, providing a more comprehensive feature representation.
Extracting modality-specific features with MLPB can increase computational cost. As shown in Tabs. \ref{MMWHS_tab} and \ref{CHAOS_tab}, while MLPB minimally affects the number of parameters, it significantly increases FLOPs. 
Future research could focus on reducing the computational cost of MLPB, making the proposed method more suitable for large-scale and real-time applications. This architecture could also be applied to other medical tasks, such as detection and classification. 
Collecting MM medical image data is inherently challenging due to privacy concerns and the need for specialized expertise, limiting access to large, diverse datasets. As a result, our current dataset includes at most four modalities.
We are committed to extending this work to scenarios with more modalities in the future.

\bibliographystyle{ieeetr}
\bibliography{refs}

\begin{thebibliography}{10}

\bibitem{douqi}
Q.~Dou {\em et~al.}, ``Unpaired multi-modal segmentation via knowledge distillation,'' {\em IEEE Transactions on Medical Imaging}, vol.~39, no.~7, pp.~2415--2425, 2020.

\bibitem{MASS}
X.~Chen {\em et~al.}, ``Mass: Modality-collaborative semi-supervised segmentation by exploiting cross-modal consistency from unpaired ct and mri images,'' {\em Medical Image Analysis}, vol.~80, p.~102506, 2022.

\bibitem{semi-CML}
S.~Zhang {\em et~al.}, ``Multi-modal contrastive mutual learning and pseudo-label re-learning for semi-supervised medical image segmentation,'' {\em Medical Image Analysis}, vol.~83, p.~102656, 2023.

\bibitem{SUMML}
L.~Zhu {\em et~al.}, ``Semi-supervised unpaired multi-modal learning for label-efficient medical image segmentation,'' in {\em Proceedings of the Medical Image Computing and Computer Assisted Intervention (MICCAI)}, pp.~394--404, 2021.

\bibitem{MeanTeacher}
A.~Tarvainen and H.~Valpola, ``Mean teachers are better role models: Weight-averaged consistency targets improve semi-supervised deep learning results,'' in {\em Proceedings of the Advances in Neural Information Processing Systems (NIPS)}, vol.~30, 2017.

\bibitem{RAM-DSIR}
Z.~Zhou {\em et~al.}, ``Generalizable medical image segmentation via random amplitude mixup and domain-specific image restoration,'' in {\em Proceedings of the European Conference on Computer Vision (ECCV)}, pp.~420--436, 2022.

\bibitem{VarDA}
F.~Wu and X.~Zhuang, ``Unsupervised domain adaptation with variational approximation for cardiac segmentation,'' {\em IEEE Transactions on Medical Imaging}, vol.~40, no.~12, pp.~3555--3567, 2021.

\bibitem{causal}
S.~Guo {\em et~al.}, ``Causal knowledge fusion for 3d cross-modality cardiac image segmentation,'' {\em Information Fusion}, vol.~99, p.~101864, 2023.

\bibitem{zhang2023multi}
S.~Zhang {\em et~al.}, ``Multi-modal contrastive mutual learning and pseudo-label re-learning for semi-supervised medical image segmentation,'' {\em Medical Image Analysis}, vol.~83, p.~102656, 2023.

\bibitem{CycleGAN}
J.-Y. Zhu {\em et~al.}, ``Unpaired image-to-image translation using cycle-consistent adversarial networks,'' in {\em Proceedings of the IEEE International Conference on Computer Vision (ICCV)}, pp.~2223--2232, 2017.

\bibitem{warp}
M.~Jaderberg {\em et~al.}, ``Spatial transformer networks,'' {\em Proceedings of the Advances in Neural Information Processing Systems (NIPS)}, vol.~28, 2015.

\bibitem{unet}
O.~Ronneberger {\em et~al.}, ``U-net: Convolutional networks for biomedical image segmentation,'' in {\em Proceedings of the Medical Image Computing and Computer-Assisted Intervention (MICCAI)}, pp.~234--241, 2015.

\bibitem{hyperfed}
Z.~Yang {\em et~al.}, ``Hypernetwork-based physics-driven personalized federated learning for ct imaging,'' {\em IEEE Transactions on Neural Networks and Learning Systems}, 2023.

\bibitem{fedBN}
X.~Li {\em et~al.}, ``Fedbn: Federated learning on non-iid features via local batch normalization,'' {\em arXiv preprint arXiv:2102.07623}, 2021.

\bibitem{PDF}
W.~Xia {\em et~al.}, ``Ct reconstruction with pdf: Parameter-dependent framework for data from multiple geometries and dose levels,'' {\em IEEE Transactions on Medical Imaging}, vol.~40, no.~11, pp.~3065--3076, 2021.

\bibitem{EMA}
S.~Laine and T.~Aila, ``Temporal ensembling for semi-supervised learning,'' {\em arXiv preprint arXiv:1610.02242}, 2016.

\bibitem{CPS}
X.~Chen {\em et~al.}, ``Semi-supervised semantic segmentation with cross pseudo supervision,'' in {\em Proceedings of the IEEE/CVF Conference on Computer Vision and Pattern Recognition (CVPR)}, pp.~2613--2622, 2021.

\bibitem{peiris2023uncertainty}
H.~Peiris {\em et~al.}, ``Uncertainty-guided dual-views for semi-supervised volumetric medical image segmentation,'' {\em Nature Machine Intelligence}, 2023.

\bibitem{valindria2018multi}
V.~V. Valindria {\em et~al.}, ``Multi-modal learning from unpaired images: Application to multi-organ segmentation in ct and mri,'' in {\em Proceedings of the 2018 IEEE Winter Conference on Applications of Computer Vision (WACV)}, pp.~547--556, 2018.

\bibitem{multi1}
D.~Nie {\em et~al.}, ``Fully convolutional networks for multi-modality isointense infant brain image segmentation,'' in {\em Proceedings of the IEEE International Symposium on Biomedical Imaging (ISBI)}, pp.~1342--1345, 2016.

\bibitem{multi2}
G.~van Tulder and M.~de~Bruijne, ``Learning cross-modality representations from multi-modal images,'' {\em IEEE Transactions on Medical Imaging}, vol.~38, no.~2, pp.~638--648, 2018.

\bibitem{pei2021disentangle}
C.~Pei {\em et~al.}, ``Disentangle domain features for cross-modality cardiac image segmentation,'' {\em Medical Image Analysis}, vol.~71, p.~102078, 2021.

\bibitem{cover1967nearest}
T.~Cover and P.~Hart, ``Nearest neighbor pattern classification,'' {\em IEEE Transactions on Information Theory}, vol.~13, no.~1, pp.~21--27, 1967.

\bibitem{knowlton1993learning}
B.~J. Knowlton and L.~R. Squire, ``The learning of categories: Parallel brain systems for item memory and category knowledge,'' {\em Science}, vol.~262, no.~5140, pp.~1747--1749, 1993.

\bibitem{rosch1973natural}
E.~H. Rosch, ``Natural categories,'' {\em Cognitive Psychology}, vol.~4, no.~3, pp.~328--350, 1973.

\bibitem{li2021adaptive}
G.~Li {\em et~al.}, ``Adaptive prototype learning and allocation for few-shot segmentation,'' in {\em Proceedings of the IEEE/CVF Conference on Computer Vision and Pattern Recognition (CVPR)}, pp.~8334--8343, 2021.

\bibitem{zhou2022rethinking}
T.~Zhou {\em et~al.}, ``Rethinking semantic segmentation: A prototype view,'' in {\em Proceedings of the IEEE/CVF Conference on Computer Vision and Pattern Recognition (CVPR)}, pp.~2582--2593, 2022.

\bibitem{zhou2022generalizable}
Z.~Zhou {\em et~al.}, ``Generalizable cross-modality medical image segmentation via style augmentation and dual normalization,'' in {\em Proceedings of the IEEE/CVF Conference on Computer Vision and Pattern Recognition (CVPR)}, pp.~20856--20865, 2022.

\bibitem{polytope}
Y.~M. Asano {\em et~al.}, ``Self-labelling via simultaneous clustering and representation learning,'' {\em arXiv preprint arXiv:1911.05371}, 2019.

\bibitem{sinkhorn}
M.~Cuturi, ``Sinkhorn distances: Lightspeed computation of optimal transport,'' {\em Proceedings of the Advances in Neural Information Processing Systems (NIPS)}, vol.~26, 2013.

\bibitem{PSI}
H.~Jeffreys, ``An invariant form for the prior probability in estimation problems,'' {\em Proceedings of the Royal Society of London. Series A. Mathematical and Physical Sciences}, vol.~186, pp.~453--461, 1946.

\bibitem{MMWHS}
``Mm-whs: Multi-modality whole heart segmentation.'' \url{https://zmiclab.github.io/zxh/0/mmwhs/}, 2023.

\bibitem{BTCV}
``Multi-atlas labeling beyond the cranial vault - workshop and challenge.'' \url{https://www.synapse.org/\#!Synapse:syn3193805/wiki/89480}, 2015.

\bibitem{CHAOS}
A.~E. Kavur {\em et~al.}, ``Chaos challenge - combined (ct-mr) healthy abdominal organ segmentation,'' {\em Medical Image Analysis}, vol.~69, p.~101950, 2021.

\bibitem{UAMT}
L.~Yu {\em et~al.}, ``Uncertainty-aware self-ensembling model for semi-supervised 3d left atrium segmentation,'' in {\em Proceedings of the International Conference on Medical Image Computing and Computer-Assisted Intervention (MICCAI)}, pp.~605--613, 2019.

\bibitem{ICT}
V.~Verma {\em et~al.}, ``Interpolation consistency training for semi-supervised learning,'' {\em Neural Networks}, vol.~145, pp.~90--106, 2022.

\bibitem{EVIL}
Y.~Chen {\em et~al.}, ``Evidence-based uncertainty-aware semi-supervised medical image segmentation,'' {\em Computers in Biology and Medicine}, vol.~170, p.~108004, 2024.

\bibitem{Adam}
D.~P. Kingma and J.~Ba, ``Adam: A method for stochastic optimization,'' {\em arXiv preprint arXiv:1412.6980}, 2014.

\bibitem{VAE}
Y.~Shi {\em et~al.}, ``Variational mixture-of-experts autoencoders for multi-modal deep generative models,'' {\em Proceedings of the Advances in Neural Information Processing Systems (NIPS)}, vol.~32, 2019.

\end{thebibliography}

\end{document}